\pdfoutput=1

\documentclass[letterpaper, 10 pt, conference]{ieeeconf}

\IEEEoverridecommandlockouts
\makeatletter\@ifundefined{iflatexml}{\newif\iflatexml}{}\makeatother
\usepackage{graphicx}
\graphicspath{{figures/}}
\usepackage{amsmath}
\usepackage{amsfonts}
\usepackage{bbm}
\usepackage{hyperref}
\usepackage{algorithm}
\usepackage{algpseudocode}
\usepackage{makecell}
\usepackage[multiple]{footmisc}
\usepackage{tabularx}
\usepackage{array}
\usepackage{longtable}
\usepackage[table]{xcolor}
\usepackage{booktabs}
\usepackage{balance}
\usepackage{needspace}

\ifdefined\pdfminorversion\pdfminorversion=4\fi

\usepackage[backend=bibtex, sorting=none, style=ieee, doi=false, url=false, eprint=false, isbn=false,maxnames=6,minnames=1]{biblatex}
\title{\LARGE \bf
OATH: Object-Agnostic Tracking for Dexterous Hands, Enabling Rapid Physics-based Retargeting
}

\iflatexml
\author{Solvin Sigurdson$^{1,2}$ \and Prasanna Sriganesh$^{1,3}$ \and Juan Borbon$^{1}$ \and Joel Burdick$^{2}$ \and Dingqi Zhang$^{1}$ \and Jiuguang Wang$^{1}$}
\else
\author{Solvin Sigurdson$^{1,2}$, Prasanna Sriganesh$^{1,3}$, Juan Borbon$^{1}$, Joel Burdick$^{2}$, Dingqi Zhang$^{1}$, and Jiuguang Wang$^{1}$%
\thanks{$^{1}$S. Sigurdson, P. Sriganesh, J. Borbon, D. Zhang, and J. Wang are with Multiply Labs, San Francisco, CA 94107, USA
        {\tt\small \{solvin, prasanna, juanb, dingqi, jw\}@multiplylabs.com}}%
\thanks{$^{2}$S. Sigurdson and J. Burdick are with the Department of Control and Dynamical Systems, California Institute of Technology, Pasadena, CA 91125, USA
        {\tt\small solvin.sigurdson@caltech.edu, jburdick@caltech.edu}}%
\thanks{$^{3}$P. Sriganesh is with the Robotics Institute, Carnegie Mellon University, Pittsburgh, PA 15213, USA
        {\tt\small pkettava@andrew.cmu.edu}}%
}
\fi

\begin{document}

\maketitle
\iflatexml
\begin{center}\small
$^{1}$S. Sigurdson, P. Sriganesh, J. Borbon, D. Zhang, and J. Wang are with Multiply Labs, San Francisco, CA 94107, USA         {\tt\small \{solvin, prasanna, juanb, dingqi, jw\}@multiplylabs.com} \\
$^{2}$S. Sigurdson and J. Burdick are with the Department of Control and Dynamical Systems, California Institute of Technology, Pasadena, CA 91125, USA         {\tt\small solvin.sigurdson@caltech.edu, jburdick@caltech.edu} \\
$^{3}$P. Sriganesh is with the Robotics Institute, Carnegie Mellon University, Pittsburgh, PA 15213, USA         {\tt\small pkettava@andrew.cmu.edu}
\end{center}
\fi

\begin{abstract}

Retargeting human demonstrations into dynamically feasible dexterous hand trajectories requires resolving complex contact physics, yet current methods either omit physical constraints or rely on computationally expensive sampling-based optimization in simulation. Reinforcement learning has previously been used to achieve planning and control through contact dynamics; however, standard policies require object-specific training and fail to generalize across arbitrary geometries. This work reframes physics-based retargeting as a residual learning problem, where we demonstrate that zero-shot generalization across objects relies on structured state representations. Specifically, formulating the observation space to include a localized contact-error signal guiding each robot link to its target coordinate on the object surface enables a single, unified residual policy to successfully grasp novel, unseen object geometries. Rolling out this policy in simulation on a kinematic reference trajectory yields a physics-based retargeter, OATH. Operating at 0.4 seconds of compute per second of reference, OATH achieves 2.4x faster-than-real-time performance and a 60–350$\times$ reduction in computational cost compared to sampling-based methods. The system recovers 58\% of held-out and 54\% of unseen-object DexYCB trajectories while maintaining a sub-2 cm tracking error.

\end{abstract}

\section{Introduction}

While dexterous hands hold the promise of general-purpose robotic manipulation, controlling them through contact remains a formidable challenge due to the high-dimensional and hybrid nature of the problem. Linearization-based methods employed for smooth nonlinear systems deteriorate around contact transitions and scale in complexity with the actuated d.o.fs~\cite{posa2014direct, aydinoglu2022admm}. Consequently, abstracting away contact dynamics via black-box simulators and Reinforcement Learning (RL) policies has become a popular paradigm to synthesize contact-rich behaviors~\cite{andrychowicz2020dactyl, handa2023dextreme}.

However, discovering these complex contact-rich behaviors from scratch using pure RL is sample-inefficient due to the vast exploration space. While RL policies can span the spectrum from high-level planning~\cite{mandi2025dexmachinafunctionalretargetingbimanual} to low-level control~\cite{chord2026learningdexterous, regrind2026minimalist}, they often require training a dedicated, computationally expensive policy per object to succeed. To guide exploration and provide an informative prior across diverse tasks, retargeted trajectories from human demonstrations are frequently used. These references serve as dense tracking signals in learning~\cite{qin2022dexmv, liu2025dextrack} or as the nominal reference signal for feedforward control~\cite{paliwal2026doasido}.
The critical bottleneck in this pipeline is the gap between kinematics and physics. Pure kinematic retargeting ignores contact dynamics, often leading to physically infeasible references. Physics-based retargeting attempts to bridge this gap by producing dynamically feasible plans that follow the reference motion via sampling-based optimization~\cite{pan2026spiderscalablephysicsinformeddexterous}. Unfortunately, these optimization methods are computationally heavy, making them slow as reference generators for training, and prohibitively expensive for real-time deployment of feed-forward reference trajectories.

\begin{figure}[t]
    \centering
    \includegraphics[width=\linewidth]{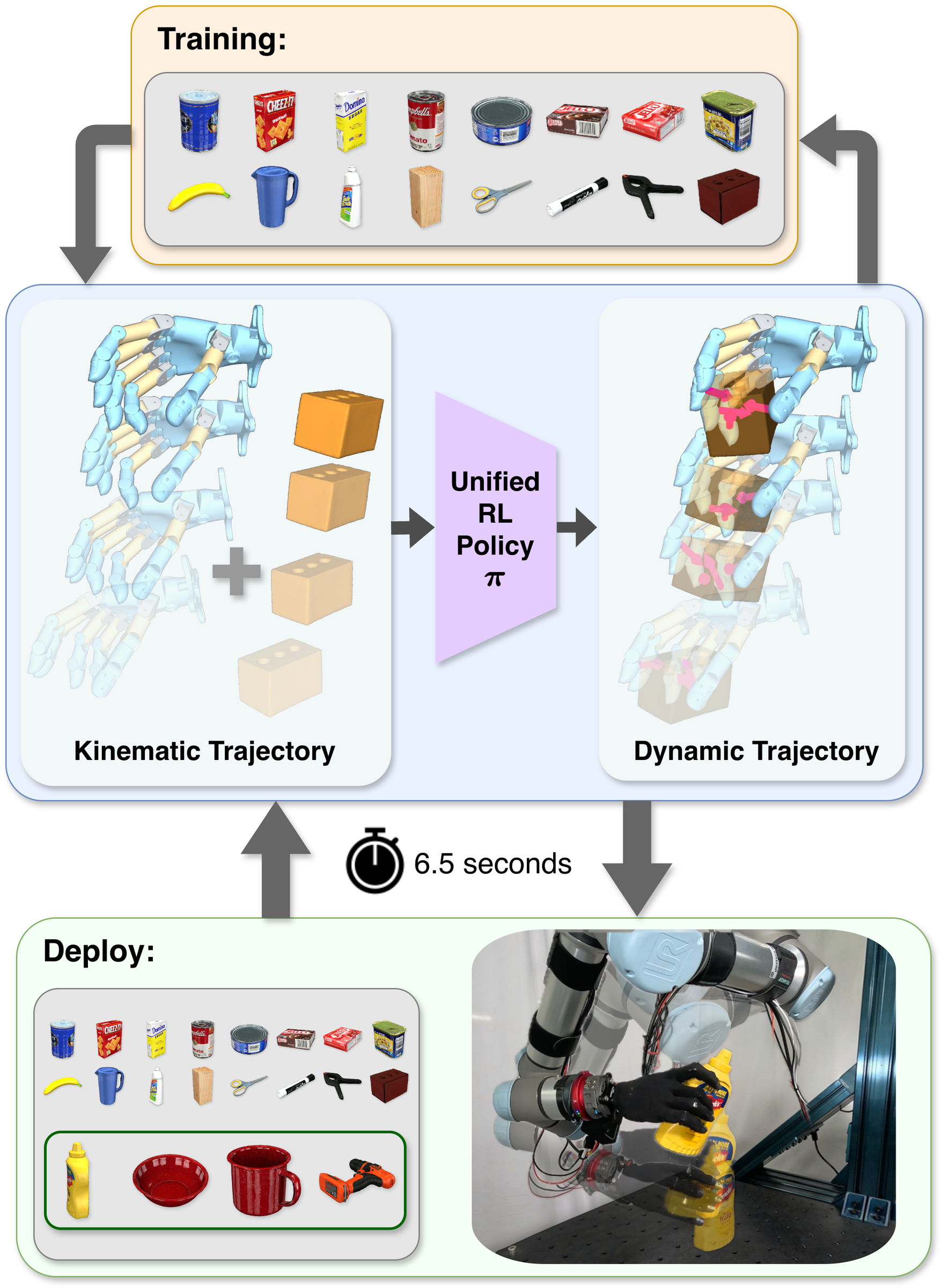}
    \par\vspace{-3pt}
    \caption{OATH trains a generalist object-agnostic controller for dexterous grasping using the DexYCB dataset. Evaluation on entirely held-out objects from the dataset demonstrates zero-shot tracking and retargeting in real time using a Wuji Hand.}
    \label{fig:overview}
    \vspace{-1.em}
\end{figure}

This work proposes a new approach (Figure~\ref{fig:overview}): by training a residual policy that focuses on contact deviation, we can successfully generalize to arbitrary objects and trajectories with a single policy. A general controller then enables online retargeting to produce dynamically feasible plans in real time, making it deployable in real-world contexts. While the method itself is generic, this paper focuses first on one of the most common types of everyday task: dexterous grasping of arbitrary object geometries for pick-and-place tasks.

\section{Background}

\textit{Learning for planning and control of dexterous hands.} Reinforcement learning (RL) provides an effective framework for contact-rich manipulation by bypassing the need to differentiate through stiff, non-smooth contact dynamics. By shifting contact exploration to physics simulation and amortizing optimal control into reactive neural policies, RL sidesteps runtime trajectory optimization entirely. This paradigm has enabled complex dynamic behaviors, ranging from single-hand reorientation~\cite{andrychowicz2020dactyl} to coordinated bimanual articulated manipulation~\cite{mandi2025dexmachinafunctionalretargetingbimanual}. 
However, in the absence of reference trajectories, RL must resolve long-horizon task planning and high-frequency closed-loop control simultaneously. This entanglement incurs severe sample complexity and yields brittle, task-specialized policies that fail to generalize across varying object geometries. Modern end-to-end visuomotor models~\cite{black2024pi0} suffer from a similar structural conflation: while internet-scale pretraining reduces downstream demonstration requirements, they remain tethered to limited task regimes and vulnerable to visual distribution shifts. Crucially, neither paradigm produces a reusable policy capable of transferring across both novel tasks \textit{and} unseen object geometries.

\textit{Reference-conditioned tracking policies.}
A descended line of work isolates control from planning by training residual policies to track a nominal reference, in the style of traditional robotics autonomy stacks. Driven by simple reward structures that penalize distance from the reference via exponential decay~\cite{peng2018deepmimic}, this paradigm now delivers real-time whole-body humanoid teleoperation and loco-manipulation on hardware~\cite{ze2025twistteleoperatedwholebodyimitation, ze2026twist2, luo2026sonicsupersizingmotiontracking}, with a single policy tracking a broad distribution of motions. This architecture structure has been carried into dexterous manipulation: CHORD~\cite{chord2026learningdexterous},  ReGrind~\cite{regrind2026minimalist}, and ManipTrans~\cite{li2025maniptrans} track retargeted references but remain limited to one object per policy. 

\textit{Retargeting human demonstrations.}
Learning dexterous skills from human video requires mapping a recorded hand--object motion onto the robot embodiment. Simple and fast retargeting methods use inverse kinematics to match demonstration fingertip positions to robot hand positions~\cite{handa2020dexpilot, qin2022dexmv}, and serve as the basis for many downstream learning works~\cite{chord2026learningdexterous, regrind2026minimalist}, but they carry only positional information: the resulting trajectories interpenetrate the object and encode nothing of the force a grasp requires. Sampling-based model predictive control~\cite{williams2017mppi, howell2022predictivesampling} can close this gap by selecting the closest motion to the kinematic reference from a distribution of simulation rollouts, and has been adopted directly as a retargeting engine~\cite{pan2026spiderscalablephysicsinformeddexterous, paliwal2026doasido} to produce dynamically feasible references through contact. The same retarget-via-controller pipeline generates whole-body humanoid data~\cite{yang2025omniretarget} and bimanual articulated manipulation~\cite{mandi2025dexmachinafunctionalretargetingbimanual}. However, their policies are trained per reference or per task family (e.g. box-picking) so each new motion still costs a training run. Sampling-based methods alone avoid training and pay at solve time instead: every clip is solved from scratch, and long-horizon sequences become prohibitively slow to generate training-time references or deploy as a feed-forward controller.

\textit{Our work.} We resolve this tension with OATH: an object-agnostic tracking policy that is grounded in contact signals extracted from the reference trajectory. Rather than relying on computationally expensive per-clip sampling-based optimization, our policy executes physics-based retargeting in seconds by amortizing dynamic feasibility into offline reinforcement learning. To guide interaction through non-smooth contact transitions, we augment nominal kinematic references with an explicit contact state representation. Restricting the policy’s objective to contact-informed tracking dramatically reduces learning complexity, enabling a single unified policy to execute dexterous grasping across novel, unseen object geometries in simulation and on hardware.

\section{Methods}

\begin{figure*}[t]
\centering
\includegraphics[width=0.8\linewidth]{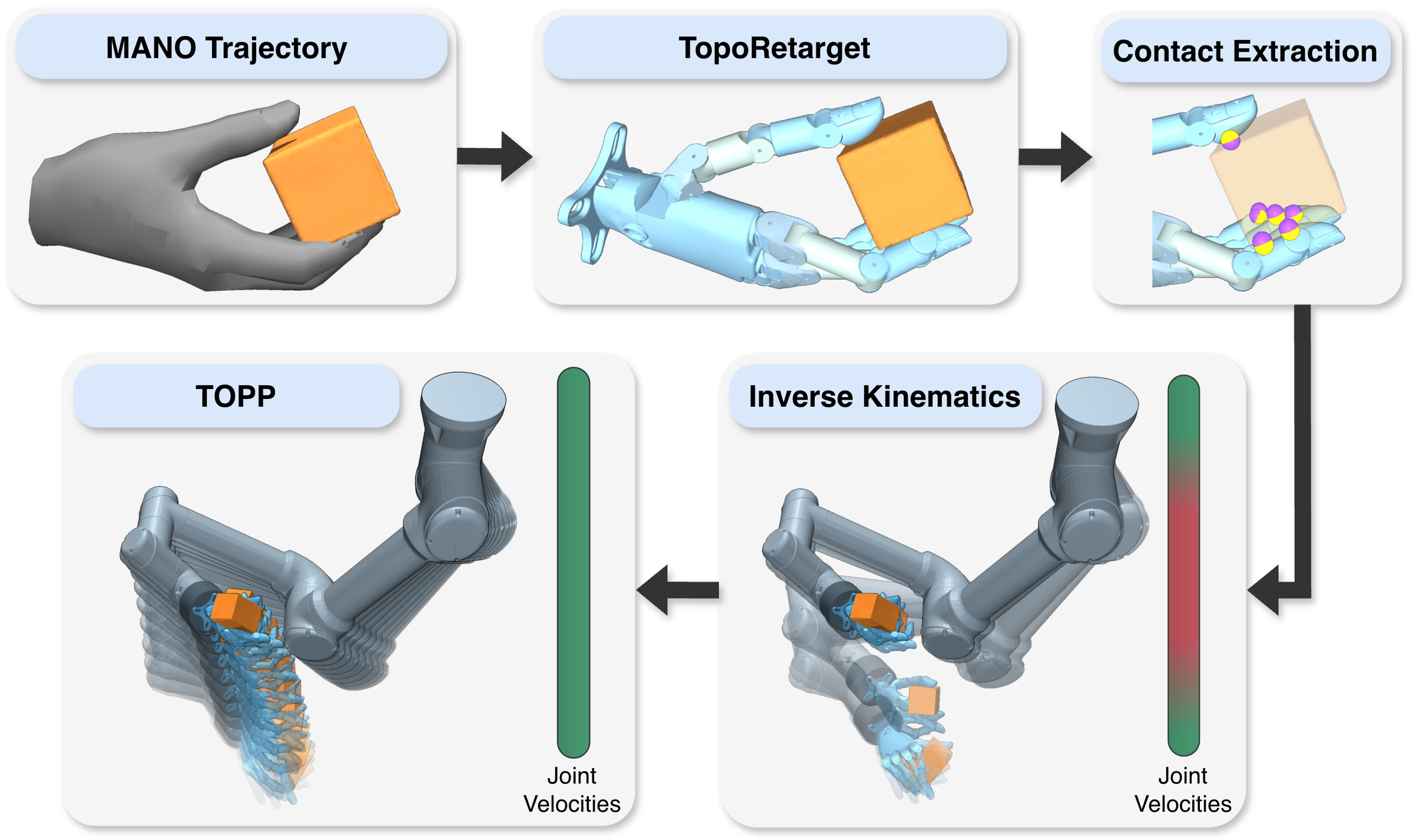}
\par\vspace{-3pt}
\caption{
\textbf{The reference generation pipeline:} (clockwise) 1) MANO meshes are fit overtop of a kinematic trajectory extracted from video or given by the DexYCB dataset, 2) TopoRetarget is applied to convert the MANO mesh to the Wuji Hand, 3) Contact locations $\bar{\mathbf{g}}_{k, i}$, $\bar{\mathbf{h}}_{k, i}$, and contact presence $\bar{m}_{k, i}$ are computed and stored for training, 4) Differential IK is applied to determine a robot arm trajectory that follows the wrist, 5) TOPP is used to slow the trajectory to a safe speed on hardware.}
\label{fig:reference-generation}
\end{figure*}

\subsection{Problem Definition}

Kinematic reference trajectories for manipulation cannot be directly executed on robotic hardware, as tracking the kinematic reference position alone is insufficient to induce contact forces. To address this, we formulate physics-based retargeting as learning a residual policy $\pi(\delta_k | \mathbf{r}_k, \mathbf{o}_k)$ operating over discrete timesteps $k \in \mathbb{N}$. Given the current reference signal $\mathbf{r}_k \in \mathbb{R}^{n_\text{ref}}$ and observation signal $\mathbf{o}_k \in \mathbb{R}^{n_\text{obs}}$, the policy predicts a residual adjustment $\delta_k \in \mathbb{R}^{n_\text{robot}}$. This residual offset is combined with the kinematic reference of the robot $\bar{\mathbf{q}}_k^\text{robot} \in \mathbb{R}^{n_\text{robot}}$ to compute the target joint command $\mathbf{u}_k$, which is tracked on the robot via PD control:
\begin{align}
\mathbf{u}_k = \bar{\mathbf{q}}_k^\text{robot} + \delta_k
\label{eq:control_input_calculation}
\end{align}
Rolling out the policy in simulation for a given kinematic reference trajectory $\bar{\mathbf{q}}_k$ produces $\mathbf{q}_k^*$, the dynamically retargeted trajectory. 
Note the kinematic reference trajectory $\bar{\mathbf{q}}_k$ is included in $\mathbf{r}_k$, although they are distinct as $\mathbf{r}_k$ includes additional task and contact information. Similarly $\mathbf{q}_k = [\mathbf{q}^\text{robot}_k, \mathbf{q}^\text{obj}_k]^T\in \mathbb{R}^n$ is a component of $\mathbf{o}_k$. Barred elements (e.g. $\bar{\mathbf{q}}_k$) refer to subsets of the kinematic reference, while unbarred elements (e.g. $\mathbf{q}_k$) are the true state.

While this residual retargeting formulation is applicable to arbitrary manipulation tasks, we focus on multi-finger dexterous grasping of diverse object geometries from human demonstration trajectories.

\begin{table*}[t]
\centering
\caption{Network, training, simulation, and reference hyperparameters}
\label{tab:hyperparameters}
\small
\setlength{\tabcolsep}{4pt}
\begin{tabular}[t]{@{}>{\raggedright\arraybackslash}p{2.9cm}@{\hspace{5pt}}>{\raggedright\arraybackslash}p{5.6cm}@{}}
\toprule
\multicolumn{2}{@{}l}{\textit{Network}} \\[1pt]
Input / action dims & 497 (state 80, reference 417) / 26 (6 arm + 20 hand) \\
MLP hidden dims & $(512, 256, 128)$, ELU \\
Encoders & separate actor / critic \\
\midrule
\multicolumn{2}{@{}l}{\textit{Optimization}} \\[1pt]
Algorithm & PPO: clip 0.2, clipped value loss (coef.\ 1.0), GAE $\gamma$ 0.99 / $\lambda$ 0.95 \\
Optimizer & Adam, lr $10^{-3}$, KL-adaptive: $\div 1.5$ if KL $> 0.02$, $\times 1.5$ if KL $< 0.005$, clamped to $[10^{-5}, 10^{-2}]$ \\
Batch per iteration & 24 steps $\times$ 16{,}384 envs (4096 per GPU $\times$ 4) = 393k transitions; 5 epochs, 4 minibatches \\
Entropy coef.\ / grad-norm clip & 0.008 / 1.0 \\
Obs.\ normalization & running mean / std (actor and critic) \\
Iterations & 15k ($\approx$5.9 B env steps, $\sim$10 h on 4$\times$ L40S) \\
\midrule
\multicolumn{2}{@{}l}{\textit{Environment and simulation}} \\[1pt]
Physics / control & 1/120 s step, 4 substeps per 30 Hz policy step \\
Solver & Newton, 20 iter., 30 line-search; implicit-fast integrator; elliptic cone, impratio 10 \\
Contacts & object condim 4, solref (15 ms, 1.0), solimp (0.85, 0.85, 0.001, 0.5, 2.0); hand condim 3; margin 0 \\
Low-level control & fingers PD ($k_p$ 0.18--0.69 N$\cdot$m/rad); arm PID + feed-forward ($k_p$ 80--1000 N$\cdot$m/rad, $k_d = 0.2\,k_p$) \\
Residual band & fingers 0.25 $\times$ joint travel ($\pm$0.35--0.79 rad); arm scale 0.0625 ($\pm$0.04--0.13 rad); offset clipped at 4$\times$ band \\
\bottomrule
\end{tabular}
\hfill
\begin{tabular}[t]{@{}>{\raggedright\arraybackslash}p{2.9cm}@{\hspace{5pt}}>{\raggedright\arraybackslash}p{5.6cm}@{}}
\toprule
\multicolumn{2}{@{}l}{\textit{Episodes and data}} \\[1pt]
Rollout horizon $T_\text{rollout}$ & 5 s (150 steps at 30 Hz); reference state initialization \\
Initialization & clip uniform over the training set; start frame uniform over the clip (RSI); object, hand, and arm at the reference pose, zero velocity, $\delta_{k-1} = \mathbf{0}$ \\
Termination $D_\text{terminate}$ & 0.15 m object position error \\
Split $R_\text{split}$ & 80/20 train/test; 4 of 20 objects held out \\
\midrule
\multicolumn{2}{@{}l}{\textit{Domain randomization (one draw per environment)}} \\[1pt]
Object & slide friction 1.5 (fixed); torsional 0.05--0.45 (log-uniform); mass $\pm20\%$; contact time const.\ 10--30 ms, damping ratio 0.8--1.2 \\
Hand & slide friction 1.0 (fixed); PD gains $\pm20\%$; joint damping, armature $\pm20\%$; joint friction $\pm50\%$ \\
Observation noise (uniform, per step) & object pose $\pm$5 mm, $\pm$0.57$^\circ$; hand joints $\pm$0.01 rad; joint velocities $\pm$0.05 rad/s; $\delta_{k-1}$ $\pm$0.01 \\
\midrule
\multicolumn{2}{@{}l}{\textit{Contact channel}} \\[1pt]
Bodies & 16: palm + \{proximal-abd., middle, distal\} $\times$ 5 fingers, model order \\
Detector / reduction & MuJoCo collision at margin 0; per body and frame: centroid, mean normal, minimum separation \\
\midrule
\multicolumn{2}{@{}l}{\textit{Reward and reference}} \\[1pt]
Reward $\beta$ & object pos 5, object rot 10, hand 1, wrist pos 5, wrist rot 5 \\
Reward $w$ & $w_\text{rate} = 0.1$; $w_\text{arm}$ annealed $1 \rightarrow 0$ over the first 10k it. \\
Reference horizon $H$ / stride $s$ & 1.0 s / 0.1 s (10 samples) \\
\bottomrule
\end{tabular}
\vspace{-2em}
\end{table*}

\subsection{Kinematic Reference Generation}
\label{sec:reference_generation}

\enlargethispage{-\baselineskip}%
Human demonstration videos provide high-level kinematic guidance, but direct joint-space mapping alters contact locations due to morphological disparities between human and robot end-effectors. To transform raw human demonstrations into joint reference trajectories, the pipeline executes surface retargeting, kinematic alignment, and trajectory parameterization across four sequential steps (Figure~\ref{fig:reference-generation}):

\begin{enumerate}

\item \textbf{Topology-Aware Retargeting:} Morphological discrepancies between human MANO meshes and the target robotic hand prevent direct joint mapping. We utilize TopoRetarget~\cite{toporetarget2026} to map human mesh surface interactions onto the robot hand mesh while maintaining functional contact topology.

\item \textbf{Target Contact Coordinate Extraction:} Formulating contact-error signals for policy training requires reference contact locations and contact presence labels. We replay the retargeted reference kinematically in MuJoCo (hand joints and object pose pinned per frame at timestep $k$, no dynamics) and read its collision detector at zero margin, so a body is in contact if it touches or penetrates the object. Contacts are recorded for the 16 hand bodies with collidable geometry. When a body has several contacts in a frame they are reduced to the centroid of the contact points. The witness points are stored in the object frame ($\bar{\mathbf{g}}_{k,i}$) and the body frame ($\bar{\mathbf{h}}_{k,i}$), along with an active-slot Boolean $\bar{m}_{k, i} \in \{0, 1\}$. Inactive slots carry $\bar m_{k,i} = 0$ and $\bar{\mathbf{g}}_{k,i} = \bar{\mathbf{h}}_{k,i} = \mathbf{0}$.

\item \textbf{Full-Body Kinematic Alignment:} Continuous tracking of target wrist poses requires coordinated motion across an arm-hand system. We apply differential inverse kinematics over the arm kinematic chain to align the arm with the wrist.

\item \textbf{Trajectory Time-Parameterization:} Kinematic retargeting and IK produce trajectories that may violate physical motor limits. We apply Time-Optimal Path Parameterization (TOPP)~\cite{pham2018toppra} to adjust trajectory timing subject to joint velocity, acceleration, and torque constraints.

\end{enumerate}

The output of the retargeting pipeline is $\mathbf{r}_k$, containing the robot and object states $\bar{\mathbf{q}}^\text{robot}_k$ and $\bar{\mathbf{q}}^\text{obj}_k$,  contact presence $\bar{m}_{k, i}$, and contact locations in the object and robot body frames $\bar{\mathbf{g}}_{k, i}$, $\bar{\mathbf{h}}_{k, i}$, where $i \in \mathbb{N}$ iterates over each robot rigid body. Next, we use $\mathbf{r}_k$ to train the RL policy.

\subsection{Learning Structure}

\textit{Dataset Structure} - We first separate a set of objects and their references from $\mathbf{r}_k$ into an unseen-object evaluation set. Of the remaining objects, the trajectories are split into a training and testing set for each object, by ratio $R_\text{split}$.

\textit{Episode Structure} - Within each episode, we sample a random reference trajectory from the training set, and roll out the current policy for horizon $T_\text{rollout}$, or until there is object deviation beyond $D_\text{terminate}$. We follow Reference State Initialization~\cite{peng2018deepmimic} and conduct domain randomization over physical properties including friction coefficients, mass, and robot control gains. At reset, the object, hand and arm are placed at the reference pose of the sampled start frame with zero velocity and $\delta_{k-1} = \mathbf{0}$; the start frame is drawn uniformly over the whole clip, so episodes may begin mid-grasp. Physical randomization is drawn once per environment at start-up; observation noise is redrawn at every control step.

\needspace{3\baselineskip}%
\textit{Network and Action Structure} - The network largely follows~\cite{ze2025twistteleoperatedwholebodyimitation} and is a temporal convolution followed by a simple MLP, with ELU activation, mapping from input size $n_\text{ref} + n_\text{obs}$ to the robot dimension $n_\text{robot}$. The policy is trained with PPO~\cite{schulman2017ppo} using the rsl\_rl implementation~\cite{rudin2022rslrl}; actor and critic receive the same observation. 

\textit{Curriculum Structure} - The only curriculum applied is on the arm-penalty residual reward, decreasing it to zero over the first 10k iterations. This helps convergence by preventing small exploratory shoulder movements from inducing large wrist deviations early in training.

\textit{Simulation} - Training and evaluation run in MuJoCo~\cite{todorov2012mujoco} on GPU via mjlab~\cite{zakka2025mjlab}, with a 1/120 s physics step and 4 substeps per 30 Hz policy step, the Newton solver (20 iterations, 30 line-search iterations, implicit-fast integrator), and an elliptic friction cone (impratio 10). Object collision geometry uses condim 4 with nominal solref (15 ms, 1.0) and solimp (0.85, 0.85, 0.001, 0.5, 2.0); hand geometry uses condim 3 at MuJoCo defaults. The contact margin is zero, so ``contact'' means touch or penetration.

\subsection{Reward Structure}

The reward is designed to be simple and promote the single goal of the policy: track the reference signal while being agnostic to object geometry. To this end, we follow the multiplicative exponential reward proposed in~\cite{ze2025twistteleoperatedwholebodyimitation}.
\begin{align}
    R_{k, \text{track}} = \prod_{i} e^{-\beta_i d_{i, k}(\mathbf{r}_k,\ \mathbf{q}_k)}
    \label{eq:multiplicative_exponential_reward}
\end{align}
where $i$ iterates over different parts of the reference signal and $d_{i, k}$ measures the distance between the reference signal and the current state. In the following experiments, this includes 5 terms: object position, object orientation, hand joint angles, wrist position, and wrist orientation. Here $d_{i,k}$ is the Euclidean distance for positions, $\sin^2(\theta/2)$ for orientations (with $\theta$ the rotation angle between current and reference), and the per-joint RMS error for joint angles. It could flexibly include any further signals that are necessary or available in the future, such as wrench measurements and tactile feedback.

We have two additional regularizing rewards on the action rate and the arm action magnitude: 
\begin{align}
    R_{k, \text{rate}} &= -w_\text{rate} \sum_j s_j \left(\delta_{k, j} - \delta_{k-1, j}\right)^2
    \label{eq:action_rate_reward} \\
    R_{k, \text{arm}} &= -w_\text{arm} \frac{1}{n_\text{arm}} \sum_{j \in \text{arm}} \delta_{k, j}^2
    \label{eq:arm_magnitude_reward}
\end{align}
where $s_j$ normalizes each action column by its residual scale (mean 1) so the rate penalty is uniform across joints.

\subsection{Reference and Observation Structure}

The policy takes as input the kinematic reference trajectory via $\mathbf{r}_k$ and current state observations via $\mathbf{o}_k$, mapping them to $\delta_k$. The subcomponents of each input are detailed below.

\needspace{3\baselineskip}%
\vspace{0.5em}
\textit{Reference Terms} - $\mathbf{r}_k$
\vspace{0.15em}

\begin{itemize}
    \item $\bar{\mathbf{q}}^\text{obj}_{k:k+H/s}$ -- object pose over a horizon
    \item $\bar{\mathbf{q}}^\text{robot}_{k:k+H/s}$ -- robot joint angles over a horizon
    \item $\bar{\mathbf{q}}^\text{wrist}_{k:k+H/s}$ -- wrist pose over a horizon
    \item $\bar{m}_{k, i}$ -- contact presence for each robot body
    \item $\bar{\mathbf{g}}_{k, i}$, $\bar{\mathbf{h}}_{k, i}$ -- contact locations on the object and on robot body $i$, in their respective body frames
\end{itemize}
\vspace{0.5em}

Wrist pose $\bar{\mathbf{q}}^\text{wrist}_{k}$ is an auxiliary reference, controlled indirectly through arm-joint residuals rather than commanded independently. 

\begin{figure}[t]
    \centering
    \includegraphics[width=\linewidth]{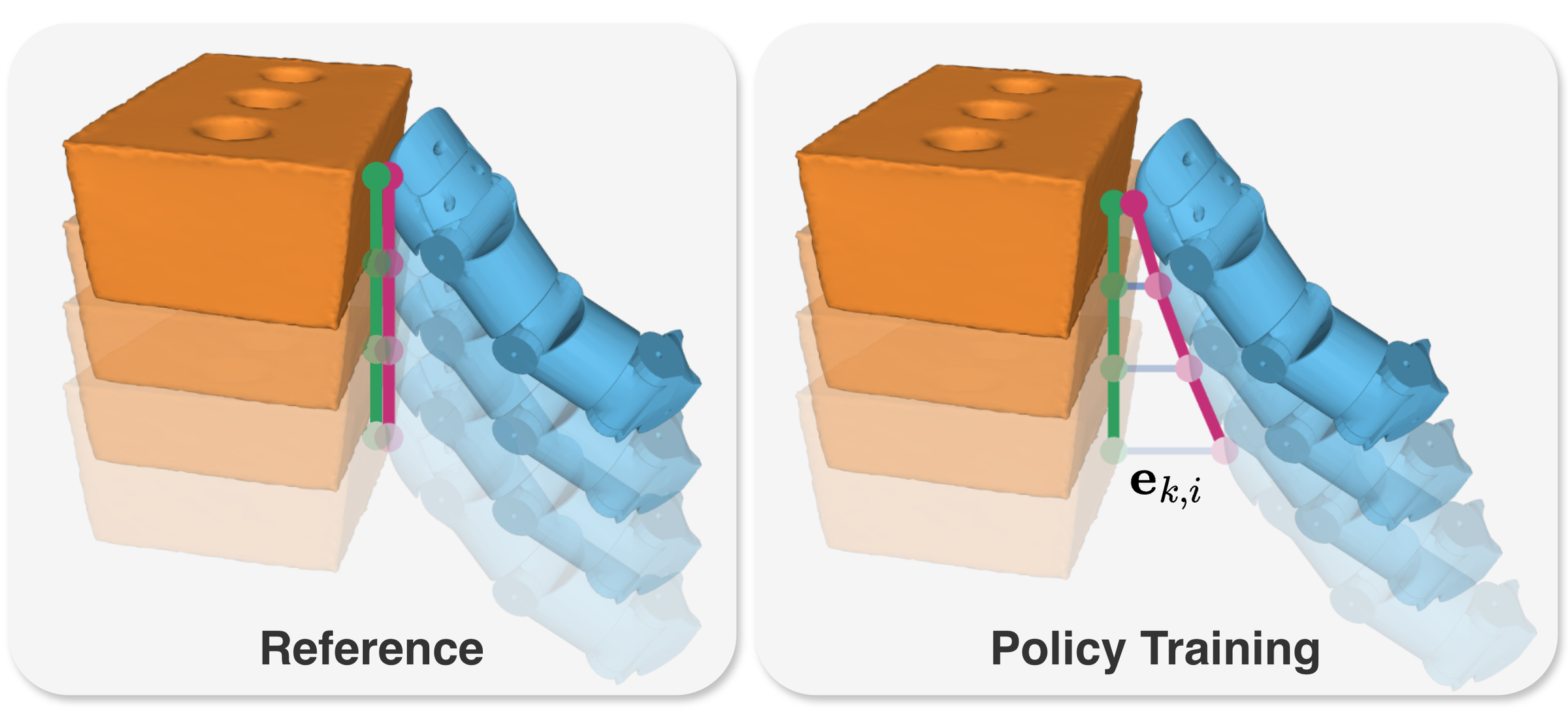}
    \par\vspace{-3pt}
    \caption{(Left) Reference contact points $\bar{\mathbf{g}}_{k, i}$, $\bar{\mathbf{h}}_{k, i}$ are extracted from the kinematic trajectory. (Right) During policy training, the deviation between the reference contact points $\mathbf{e}_{k, i}(\mathbf{q}_k)$ is computed and passed to the network as an observation.}
    \label{fig:contact-error}
    \vspace{-1.5em}
\end{figure}

\vspace{0.5em}
\textit{Observation Terms} - $\mathbf{o}_k$
\vspace{0.15em}

\begin{itemize}
    \item $\mathbf{q}_k$ -- current robot and object state 
    \item $\mathbf{q}_k^\text{wrist}$ -- current wrist state
    \item $\dot{\mathbf{q}}^\text{robot}_k$ -- robot joint velocities
    \item $\delta_{k-1}$ -- previous residual action
    \item $\mathbf{e}_{k, i}$ -- contact error vector for each robot body, Eq.~\ref{eq:contact_error}
\end{itemize}
\vspace{0.5em}

A key insight for our policy is to pass the vector $\mathbf{e}_{k, i}$ (Figure~\ref{fig:contact-error}): the current vector between contact locations from the reference. This term is a function of the current state, and serves as an error signal to guide bodies back into alignment with the reference signal contacts. This is important since the kinematic reference most strongly implies existence of a control input to track the reference if the current and reference contact states match. This is very similar to virtual contact guidance in~\cite{pan2026spiderscalablephysicsinformeddexterous}, but applied as a signal in RL.
\begin{equation}
    \mathbf{e}_{k, i} = \left[\mathbf{p}_{k, o} + \mathbf{R}_{k, o}\bar{\mathbf{g}}_{k, i}\right] -  \left[\mathbf{p}_{k, i} + \mathbf{R}_{k, i}\bar{\mathbf{h}}_{k, i}\right]
    \label{eq:contact_error}
\end{equation}
with $(\mathbf{p}_{k, o}, \mathbf{R}_{k, o}), (\mathbf{p}_{k, i}, \mathbf{R}_{k, i})$ denoting world poses of the object and of robot body $i$, both functions of $\mathbf{q}_k$.

\begin{figure}[b]
    \centering
    \includegraphics[width=0.9\linewidth]{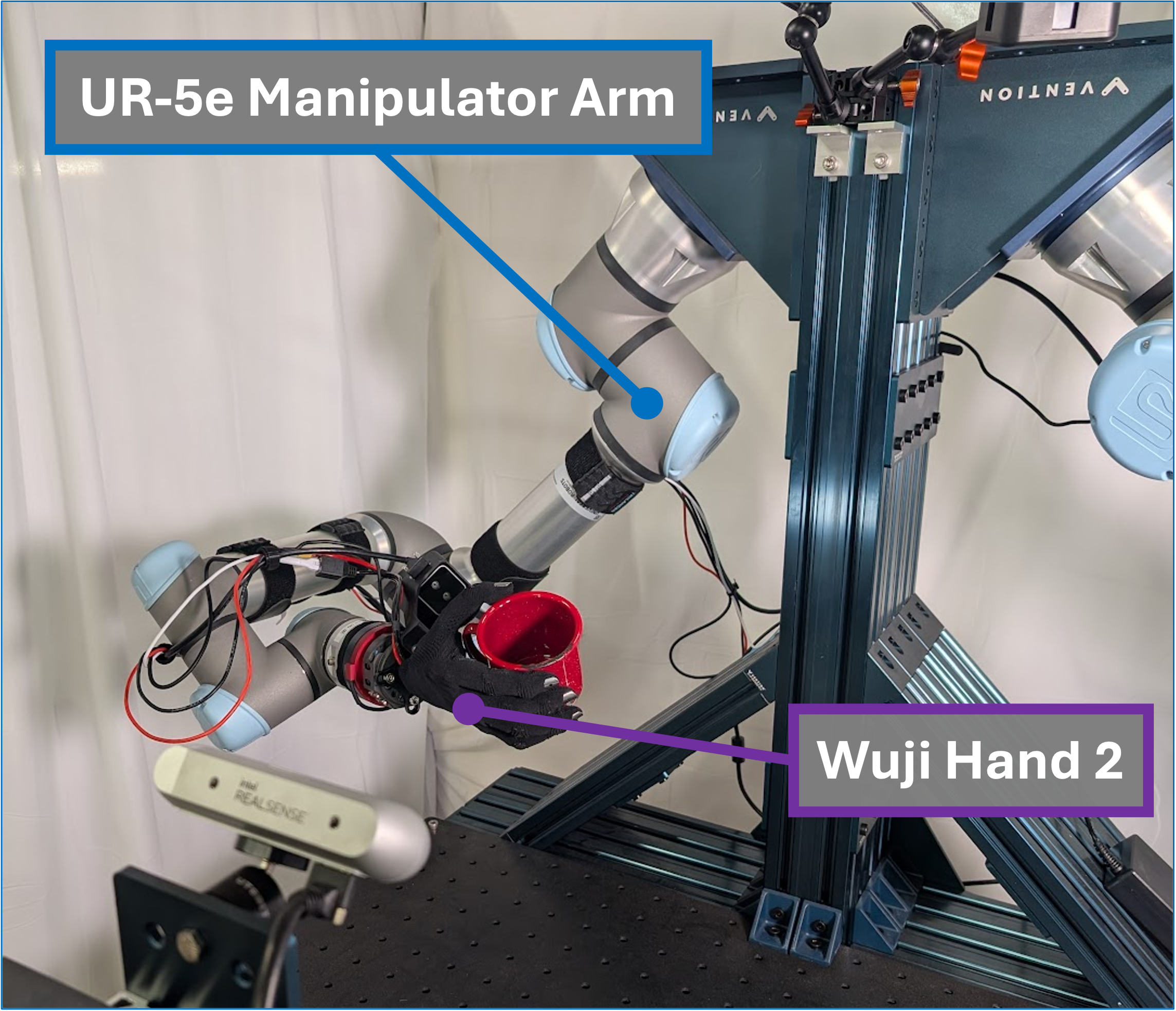}
    \par\vspace{-3pt}
    \caption{\textbf{The station:} a UR-5e and Wuji 2 hand holding the mug object from DexYCB dataset}
    \label{fig:starport-setup}
\end{figure}

\begin{table*}[t]
\centering

\caption{Controller and Baseline Performance on DexYCB Tasks}
\label{tab:grasping}

\resizebox{\linewidth}{!}{%
\begin{tabular}{lccccccc}
\toprule
 & \multicolumn{4}{c}{Baselines} & \multicolumn{3}{c}{OATH} \\
\cmidrule(lr){2-5}\cmidrule(lr){6-8}
Metric & SPIDER~\cite{pan2026spiderscalablephysicsinformeddexterous} & DAID~\cite{paliwal2026doasido} & ReGrind~\cite{regrind2026minimalist} & CHORD~\cite{chord2026learningdexterous} & Train & Test & Unseen \\
\midrule
($\uparrow$) Success rate (\%) & 22.0 & \textbf{100} & 5.4 & 2.0 & 90 & 58 & 54 \\
($\downarrow$) Object tracking error (mm) & 29.9 & 20.4 & 43.2 & \textbf{7.9} & 11.7 & 15.3 & 13.8 \\
($\downarrow$) Compute time per second of reference & 25.2 & 138 & 0.93 & 0.88 & \multicolumn{3}{c}{\rule[0.55ex]{1.3cm}{0.5pt}\;\textbf{0.40}\;\rule[0.55ex]{1.3cm}{0.5pt}} \\
($\downarrow$) Compute per successful second of reference & 124 & 138 & 15.8 & 64.9 & 0.46 & \textbf{0.69} & \textbf{0.71} \\
\bottomrule
\end{tabular}}
\vspace{-1.6em}
\end{table*}

\begin{figure*}[t]
    \centering
    \includegraphics[width=\textwidth]{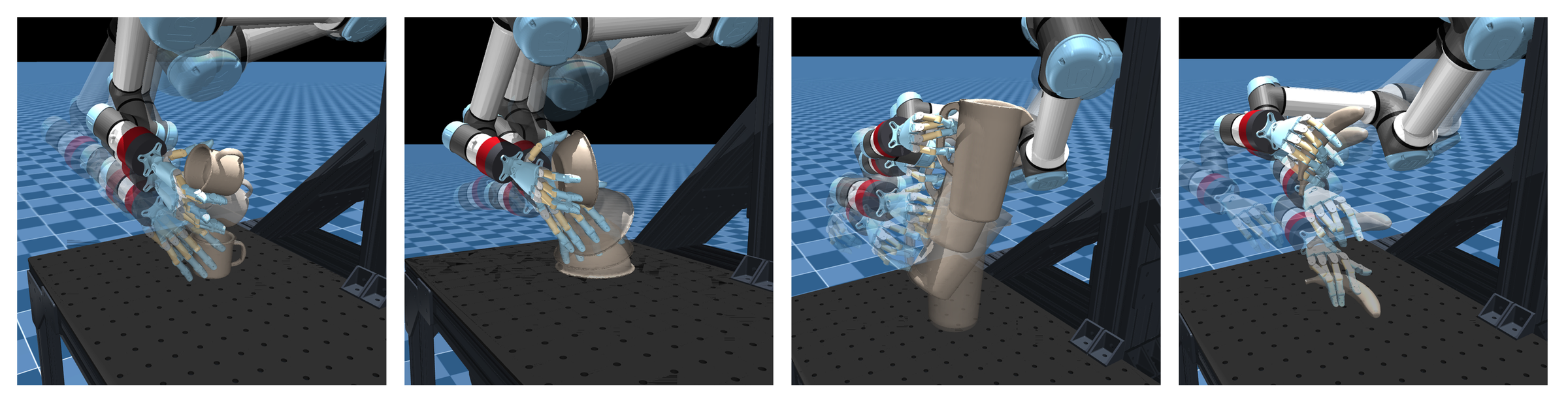}
    \par\vspace{-3pt}
    \caption{\textbf{Closed-loop policy rollouts across challenging DexYCB trajectories.} Ghosted poses illustrate motion progression, with opacity increasing over time ($t = 0 \to T$). From left to right: unseen objects—mug (narrow handle) and bowl (inverted trajectory); held-out test trajectories—pitcher (oblique contact surface) and banana (thin cross-section).}
    \label{fig:hardware-rollouts}
    \vspace{-1em}
\end{figure*}

\begin{figure*}[t]
    \centering
    \includegraphics[width=\textwidth]{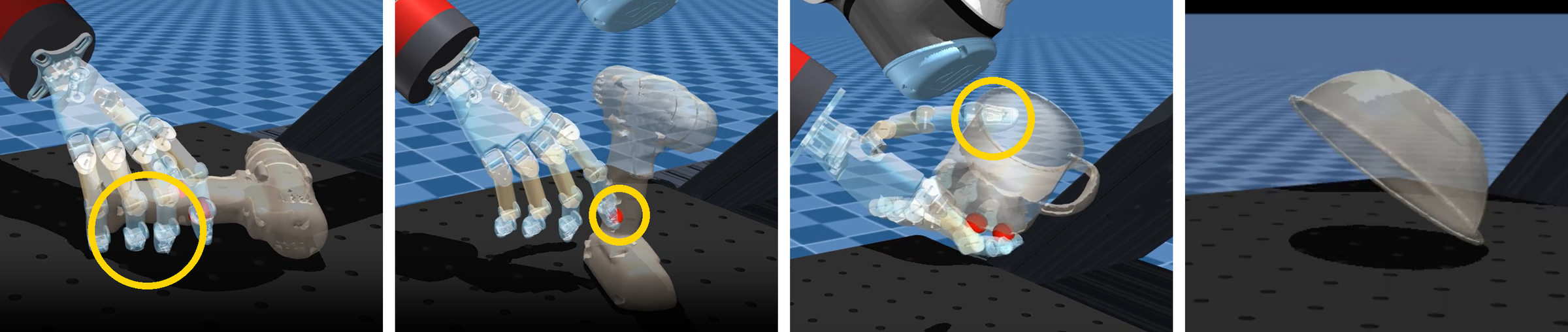}
    \par\vspace{-3pt}
    \caption{\textbf{Common failures of the kinematic retargeting:} (left to right) a) fingers penetrate the tabletop, b) spurious contact on object approach, c) missing contact references in whole prehensile grasps, d) floating object references, which fall in simulation.}
    \label{fig:kinematic-failures}
    \vspace{-0.4em}
\end{figure*}

\section{Experimental Results and Discussion}

\subsection{Dataset}

We use 340 trajectories from the DexYCB~\cite{chao2021dexycb} dataset to demonstrate dexterous grasping of diverse object shapes. We hold out 20 percent of object types as completely unseen by training: the mustard bottle, mug, bowl, and power drill. The remaining 276 trajectories get an 80/20 split into training and test trajectories respectively. Each trajectory is then passed through our kinematic retargeting pipeline described in Section~\ref{sec:reference_generation}. After TOPP is applied to the dataset, the average trajectory length is 5.82 seconds.

\subsection{Embodiment and Hardware}

We train and deploy the policies on a UR-5e robot arm (6 dofs) equipped with a Wuji 2 robot hand (20 dofs), seen in Figure~\ref{fig:starport-setup}. Policy training was conducted on 4x Nvidia L40S GPUs and policy rollouts are timed on an Nvidia RTX 5080 GPU and Intel i9-14900K CPU.

\subsection{Controller and Retargeting Performance}

We evaluate the performance of our closed-loop controller in simulation on the training/testing/unseen split.  Table~\ref{tab:grasping} shows the success rate (percentage of runs that bring the object within 10 centimeters of its goal position), object tracking error, the computation time spent per second of reference trajectory length, and the computation time spent per second of \textit{successful} reference trajectory length. Both compute metrics are normalized to RTX 5080 processing speeds, as it was not always possible to run evaluation serially on the same machine. We additionally compare to four baseline methods: SPIDER~\cite{pan2026spiderscalablephysicsinformeddexterous} and DAID~\cite{paliwal2026doasido} (sampling-based controllers), and ReGrind~\cite{regrind2026minimalist} and CHORD~\cite{chord2026learningdexterous} (single-task RL-based controllers). The baselines are evaluated on 50 randomly selected kinematic reference trajectories from our dataset, except for DAID, which does its own kinematic retargeting. Our policy takes $\sim$10 hours to train and can then be deployed immediately without any retraining or fine-tuning. Demonstrations of the policy successfully grasping challenging unseen trajectories and objects involving thin surfaces and inverted motions can be seen in Figure~\ref{fig:hardware-rollouts}.

The relatively high success rate compared to all non-DAID baselines and low tracking error in all columns of Table~\ref{tab:grasping} indicate the ability of our control policy to rapidly convert kinematic references to dynamically feasible trajectories. In particular, the success on unseen objects demonstrates its object-agnostic abilities.

\begin{figure*}[t]
    \centering
    \includegraphics[width=\textwidth]{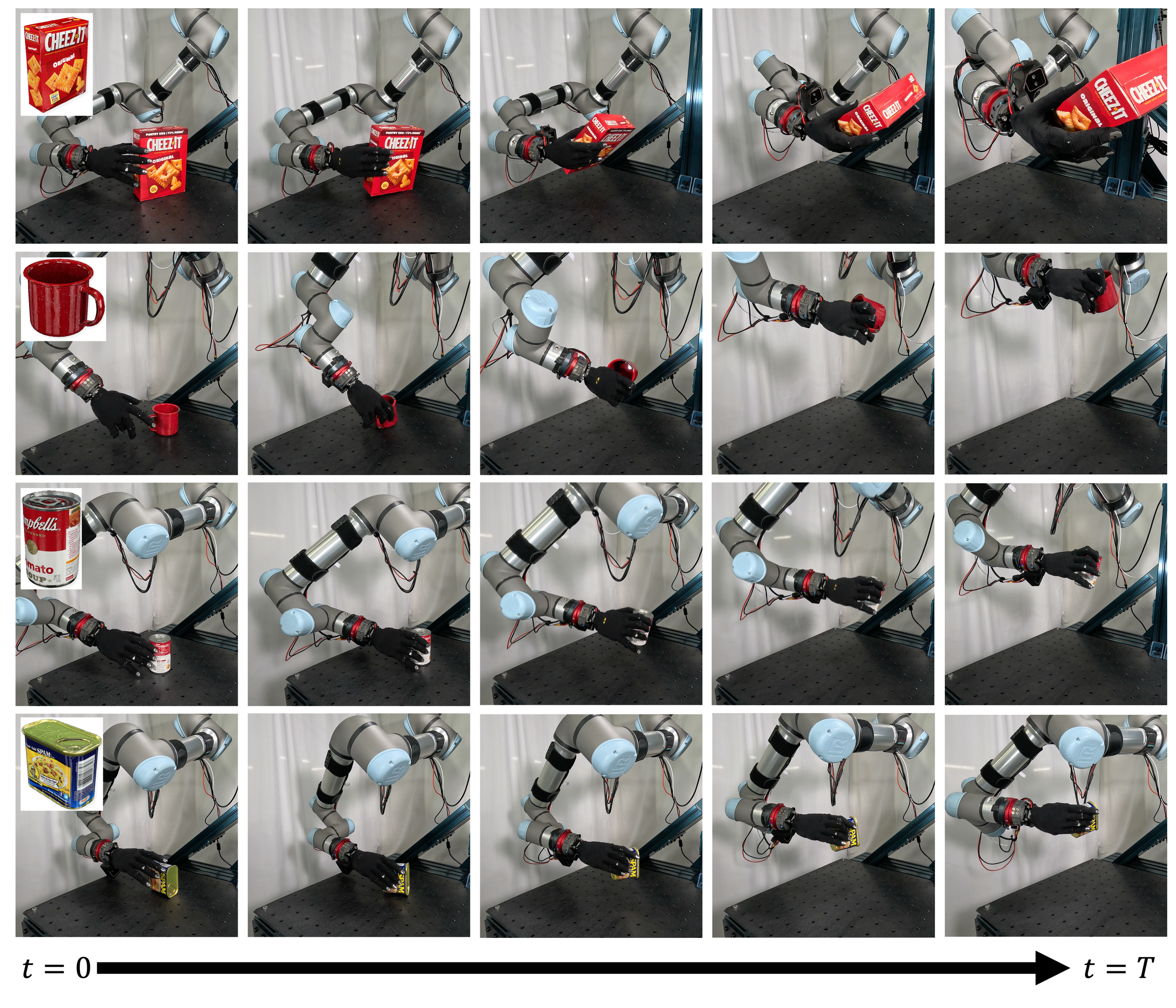}
    \par\vspace{-3pt}
    \caption{\textbf{Open-loop hardware execution on held-out DexYCB trajectories}. Sequential stills ($t=0 \to T$) of open-loop joint playback on a UR5e manipulator equipped with a Wuji Hand 2. All four evaluations follow held-out trajectory references, with the mug testing an entirely unseen object geometry. When compounding tracking errors remain bounded, open-loop execution successfully manipulates the objects, demonstrating the physical plausibility of the retargeted references.}
    \label{fig:real-robot-results}
    \vspace{-1em}
\end{figure*}

\begin{table*}[t]
\centering
\caption{Reference and Observation Ablations}
\label{tab:ablation}
\resizebox{\linewidth}{!}{%
\begin{tabular}{lcccccc}
\toprule
 & \multicolumn{2}{c}{Training} & \multicolumn{2}{c}{Testing} & \multicolumn{2}{c}{Unseen} \\
\cmidrule(lr){2-3}\cmidrule(lr){4-5}\cmidrule(lr){6-7}
Reference/Observation input & SR (\%) & Err. (mm) & SR (\%) & Err. (mm) & SR (\%) & Err. (mm) \\
\midrule
All terms: $\bar{\mathbf{q}}^\text{obj}_{k:k+H},\ \bar{\mathbf{q}}^\text{robot}_{k:k+H},\ \bar{m}_{k,i}, \mathbf{e}_{k,i}$ & 89 & 11.6 & 53 & 14.7 & 46 & 14.9 \\
Remove $\bar{m}_{k,i}, \mathbf{e}_{k,i}$ & 69 & 16.1 & 31 & 17.4 & 38 & 14.5 \\
Remove $\bar{\mathbf{q}}^\text{robot}_{k:k+H}$ & 89 & 13.1 & 47 & 19.4 & 52 & 16.2 \\
Replace $\bar{\mathbf{q}}^\text{robot}_{k:k+H}$ with $\bar{\mathbf{q}}^\text{robot}_{k}$ & 89 & 12.0 & 56 & 14.9 & 52 & 13.1 \\
Replace $\bar{\mathbf{q}}^\text{obj}_{k:k+H}$ with $\bar{\mathbf{q}}^\text{obj}_{k}$ & 90 & 11.7 & 58 & 15.3 & 54 & 13.8 \\
\bottomrule
\end{tabular}}
\vspace{-2em}
\end{table*}

The baselines' performance helps reveal the advantages of our controller. Sampling-based methods have the potential to be highly performant - as demonstrated by DAID's 100\% success rate - but remain unable to transfer to real-time deployment due to their long computation time. In contrast, the single-task RL methods are fast but unable to generalize strongly beyond the single clip they trained on. Our method sits in the middle, with reasonable performance across a variety of tasks along with faster than real-time deployability. This is effectively summarized in the final row of the table, in which our method surpasses all others by 20--200$\times$. On success rate alone, our method surpasses all others when given only kinematically retargeted references, except for DAID, which is unsurprising given it has $\sim345\times$ more compute available.

\textit{Unseen Object Failures.} Investigating failures on unseen objects more closely, we find that many stem from poor kinematic retargeting provided as a reference to the controller. Examples of these failures can be seen in Figure~\ref{fig:kinematic-failures}. This implies that when our controller is given a physically meaningful grasp, it can generate the necessary forces to move the object. As future efforts improve kinematic references, our method can readily fill in the required actuation signals.

\textit{Hardware deployment.} To evaluate the dynamic plausibility of our retargeted motions on physical systems, we execute the synthesized trajectories open-loop on the hardware system. Because open-loop playback lacks sensory feedback to recover from real-world disturbances, unmodeled contact slip and small calibration offsets can compound into task failures. However, when compounding drift remains bounded, the generated trajectories successfully coordinate multi-fingered manipulation directly on hardware. As illustrated in Fig.~\ref{fig:real-robot-results}, we evaluate across different objects from the DexYCB dataset using held-out reference trajectories, where the mug additionally serves as an entirely unseen object geometry. While closed-loop tactile or visual feedback remains necessary for high success rates in contact-rich settings, these hardware trials demonstrate that our retargeting pipeline generates dynamically plausible references capable of transferring to real multi-fingered systems.

\textit{Ablations.} Next, we ablate different parts of our reference signal in Table~\ref{tab:ablation} in order to determine which references are important for the tracking controller. Each row indicates what factors are removed from the standard reference signal, and the columns indicate the Success Rate and tracking error on different parts of the dataset. 

Table~\ref{tab:ablation} shows that the contact reference dominates: removing $\bar m_{k,i}$ and $\mathbf e_{k,i}$ lowers success by 21, 27 and 16 points on the training, test and unseen splits respectively, consistent with contact presence and location being what determines whether the forces of the demonstration can be reproduced. The remaining terms carry little information. Replacing the object window with the current object pose does not reduce performance and slightly improves the held-out splits, and the robot-pose window is likewise dispensable, suggesting that the current object error already specifies the tracking objective and that additional lookahead is redundant rather than informative. Note that the best reference structure results are reported against the baselines in Table~\ref{tab:grasping}.

\textit{Real-World Viability.} Success on a variety of objects from DexYCB demonstrates the utility of our method: household tasks are often composed of repeated pick-and-place trajectories of everyday objects. Given a single demonstration, we obtain a dynamically feasible trajectory within seconds, unlocking practical everyday use. On average the kinematic retargeting pipeline takes 4 seconds across DexYCB clips, and policy rollout takes another 2.3 seconds, leading to a total retargeting time of $\sim6.5$s per clip.

We attribute the success and real-world potential of our method to two main factors. First, with the contact signal provided as a reference, we can effectively generalize to arbitrary object geometries, allowing more versatile performance than CHORD and ReGrind~\cite{chord2026learningdexterous, regrind2026minimalist}.
Second, by enforcing dynamic feasibility via a single policy rollout instead of hundreds to thousands as in sampling-based optimization and RL, we gain the necessary efficiency to scale to long-horizon tasks and real-world deployment.

\section{Conclusion and Future Work}

By leveraging contact-based observation signals and training over a diverse set of object geometries, this work demonstrates the feasibility of OATH, an object-agnostic general RL policy. OATH enables physics-based retargeting of dexterous grasping on completely unseen objects in the DexYCB dataset, validated in simulation and deployed open-loop on hardware. Retargeting time per second of successful trajectories was 20--200$\times$ lower than that of baseline methods, opening the possibility of real-time retargeting for household deployment. Future work will focus on improving the quality of the kinematic retargeting to improve the incoming signal to the physics-based retargeter, and generalizing to more dynamic contact-rich motions.

\balance
\printbibliography

\end{document}